# Efficiently Training On-Policy Actor-Critic Networks in Robotic Deep Reinforcement Learning with Demonstration-like Sampled Exploration

Zhaorun CHEN
School of Electronic Information and Electrical Engineering
Shanghai Jiao Tong University
billchan226@sjtu.edu.cn

Binhao CHEN, Shenghan XIE, Liang GONG*, Chengliang LIU
School of Mechanical Engineering
Shanghai Jiao Tong University
* gongliang_mi@sjtu.edu.cn

Zhengfeng ZHANG, Junping ZHANG
Shanghai Key Lab of Intelligent Information Processing
Fudan University

*Abstract*—In complex environments with high dimension, training a reinforcement learning (RL) model from scratch often suffers from lengthy and tedious collection of agent-environment interactions. Instead, leveraging expert demonstration to guide RL agent can boost sample efficiency and improve final convergence. In order to better integrate expert prior with on-policy RL models, we propose a generic framework for Learning from Demonstration (LfD) based on actor-critic algorithms. Technically, we first employ K-Means clustering to evaluate the similarity of sampled exploration with demonstration data. Then we increase the likelihood of actions in similar frames by modifying the gradient update strategy to leverage demonstration. We conduct experiments on 4 standard benchmark environments in Mujoco and 2 self-designed robotic environments. Results show that, under certain condition, our algorithm can improve sample efficiency by 20% ~ 40%. By combining our framework with on-policy algorithms, RL models can accelerate convergence and obtain better final mean episode rewards especially in complex robotic context where interactions are expensive.

## I. INTRODUCTION

With the rapid progress of reinforcement learning (RL), deep learning is successfully integrated in solving complex stochastic non-linear control problems in dynamic environments. The actor-critic (AC) framework is the most prevailing RL algorithm which aims at solving high dimensional problems with continuous action space. The AC framework, which combines policy gradient and value function method, skillfully tackles the trade-off where value function method suffers from the curse of dimensionality and that policy gradient approach is poor in sample efficiency and slow in convergence. However, most AC algorithms such as Proximal Policy Optimization (PPO) [1] and Asynchronous Advantage Actor-Critic (A3C) [2] still heavily rely on agent-environment interaction to improve their performance. Most severely, RL algorithms are toughly challenged by the cold start problem, which refers to the large amount of exploration that is essential for finding the appropriate policy during early training stage, which is time expensive and unnecessary. This is mainly due to the fact that the agent has completely no prior knowledge about the environment and heavily relies on autonomous exploring, within the scheme of model-free RL [3]. Fortunately, many recent research [4], [5] showed that passing the human prior and expert demonstration on the RL agent could be the answer

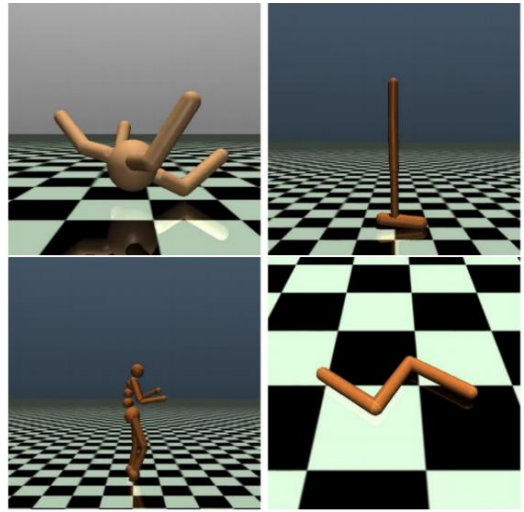

Figure 1  2D and 3D robot models used as benchmark environments: Ant-v2, Hopper-v2, Humanoid-v2, Swimmer-v2.

This work was supported by National Natural Science Foundation of China (Grant No.51775333), the Science Foundation of Shanghai Municipal Commission of Science and Technology (Grant No.18391901000) as well as the National Key Research and Development Project (2018YFB1305104).

to the cold-start dilemma, by guiding the agent to accelerate training.

As a matter of fact, in terms of the off-policy AC algorithms such as Soft Actor-Critic (SAC) [6] and Deep Deterministic Policy Gradient (DDPG) [7], it is rather simple to internalize prior knowledge by directly adding demonstration to the replay buffer. On the other hand, the integration of demonstration is often restricted to pre-training for on-policy algorithms, ultimately resulting from the absence of important sampling technique. However, it is far from enough to simply feed the agent with demonstration in pre-training, whose presence is also necessary during the formal training for the agent to fully distill the expert prior. As a result, this approach is low in efficiency and still can not deal with the cold start dilemma. Furthermore, inappropriate exploitation of demonstration data could end up in over-fitting. Therefore, it is imperative to find an effective and efficient approach to leverage the demonstration.

In this study, we propose a generic learning from demonstration (LfD) [8] framework for all actor-critic as well as policy gradient methods, which can improve convergence rate and tackles the cold start problem. Thanks to the indirect perspective of leveraging the demonstration, our approach can maintain the guidance of expert prior throughout the process of agent-environment interaction, apparently to the advantage of the on-policy algorithms. Since we usually employ gradient descent to update, we can figuratively explain it as increasing the likelihood of the promising actions and decreasing those that are below average, in terms of value [9]. Therefore, by increasing the step length of the likelihood of actions that are similar to demonstration, promising actions that are inferred from expert knowledge are more likely to be selected, thus accelerating the convergence. Technically, we employ the K-means clustering [10] to evaluate the similarity of real-time sampled experiences and demonstration data. Besides, in order to avoid over-fitting, this process will be cut out only after several epochs, while for the rest of the training, the agent is free to explore, until the final convergence.

We aim at combining human demonstration with reinforcement learning to solve tasks in diverse dynamic environments. Therefore, we firstly verify our LfD framework on standard simulation benchmark environments in Mujoco [11] as depicted in Fig. 1. Then we deploy our algorithm on a simulated humanoid robot and a mechanical robot arm developed by our lab, as is shown in Fig. 4, which are more complex but in accordance with the real application scenario. The main contributions of our work can be summarized as:

- We propose a generic LfD framework for all actor-critic algorithm, which is especially beneficial for on-policy models.
- By integrating the demonstration data, the cold start problem is tackled and the episode rewards during early stage of training are raised.
- The model can gain better sample efficiency and obtain better final performance.

II. BACKGROUND

A. Actor-critic framework

Since actor-critic framework [9] combines policy gradient and value function method, it can thus be leveraged to solve high dimensional problems with continuous action space. Structurally, it consists of an actor, usually parameterized by a neural network, which generates an action at given states, and a critic, which serves to estimate the value of the selected action or state. Thanks to the presence of the critic network, the actor network can directly update with estimation of accumulated rewards, saving from having to collect sufficient interactions for the values to converge, which greatly improves sample efficiency and convergence rate. Based on whether the behavioral policy is the same as the target policy, the state-of-the-art AC algorithms can be categorized into on-policy algorithms: PPO [1], A3C [2]; and off-policy algorithms: SAC [6], DDPG [7].

B. Learning from Demonstration

Within the scheme of model-free RL, the agent mainly relies on collecting a large amount of experiences from exploring the environment, and adjust its policy according to the rewards received [3]. Nevertheless, the reward is often sparse in high dimensional observation space, resulting in many random and inefficient attempts of exploration [12], especially during early training stage. However, considering that when learning a new skill, human beings usually begin with learning from their instructors and then trial on their own, we can apply the idea of Learning from Demonstration to RL scheme. With the guidance of demonstration, the robots can distill expert prior [5], accelerating learning and improving performance. The prevailing LfD methods can be classed into 3 categories, namely *imitation learning* [13] in a supervised way, *soft combination with RL* and *inverse reinforcement learning* [14] to obtain better design of rewards. [15] and [16] employ supervised learning approaches to fit the agents' actions to demonstration data. Moreover, [17] also proposes to train a mechanical prostheses agent within AC framework to imitate expert. However, such supervised methods leave out exploration, the most significant property in RL, which leads to a dilemma where that agents can hardly exhibit performances that are better than the expert. As is discussed before, pretraining and storing demonstration data in replay buffer are the two mainstream methods to softly combine human prior with RL. [4] handcrafts a specific environment and reward to pretrain the policy network in PPO model. However, such approach can not maintain the expert's guidance throughout the training, thus restricting the effect. On the other hand, [18] and [19] both try to leverage demonstration data by engaging them in replay buffer, respectively with a DQN and DDPG model. Nevertheless, while the convergence rate is raised, the demand for demonstration data is raised as well, and it is prone to overfit otherwise. Distinctively, [21] and [22] redesign the loss function to evaluate the similarity between demonstration data and sampled explorations, in the hope that agents could imitate the expert's actions within similar observations. However, due to the curse of dimensionality [20], the sparsity of demonstration data and the intrinsic flaw of their distance metric disable their approach from functioning in high dimensional observation space.

Our LfD pipeline employs K-Means clustering to tackle the constraint of dimensionality [23], which seeks to make most of the demonstration dataset. Leaving the loss function intact, we guide the agent to imitate experts' behaviors by adjusting the direction of gradient descent. The detailed approach will be explained in the next section.

## III. METHODS AND MATERIALS

### A. Data Preparation

Two datasets, namely expert demonstration $D$ and agent-environment interactions $E$ are present in our framework. The demonstration is usually derived by replaying a well-trained RL model, and interactions with the environments are collected by real-time exploration. They are both in $\{s_t, s_{t+1}, r_t, a_t\}$ form. To settle the problem that demonstration data are sparse and underrepresented in high dimensional space, we propose a novel metric with K-Means clustering to evaluate the similarity between the two datasets. K-means clustering classifies unlabeled data into $K$ categories in an unsupervised manner. We select the *Euclidean Distance* as our evaluation metric:

$$d_{DE} = \sqrt{(x_1 - x_2)^2 + (y_1 - y_2)^2} \quad (1)$$

By employing the K-Means pipeline, we can easily obtain the solution to the NP hard problem:

$$argmin_{C_i} E = \sum_{i=1}^{k} \sum_{x \in C_i} ||x - \mu_i||_2^2 \quad (2)$$

where $\mu_i$ is the mass center of cluster $C_i$: $\mu_i = \frac{1}{|C_i|} \sum_{x \in C_i} x$.

We first cluster the observation space and then action space on dataset $E$, which is normalised to improve generalization. Then we traverse each demonstration frame $\{s_t^D, s_{t+1}^D, r_t^D, a_t^D\}$, and select those whose distance to the nearest cluster of both observation and action are within a pre-defined threshold $H$. These selected frames are restored in a new dataset $D_p$, which consists of the promising actions similar to demonstration. Later we merge $D_p$ with $E$, and perform gradient descent to update.

### B. Demonstration Guidance

Taking account of the prior knowledge, we can modify the gradient direction to update parameters by taking a bigger step along those actions that are similar to the demonstration. As is shown in Fig. 2, it is a probability density function (PDF) of a binary Gaussian distribution with three maximum and minimum points, which can help justify the properties of our algorithm. By simplifying the parameters of the policy network to a 2D plane and representing the action likelihood with z-axis, we can derive the 3D geometric graph of action likelihood distribution with respect to network parameters. The purple line represents the traditional gradient descent method and the red line represents our method, which is guided by demonstration. Starting from the same initial state, the traditional method naturally traps in a local optima, while our algorithm can converge near the global optima, stepping further along the promising direction that demonstration recommends.

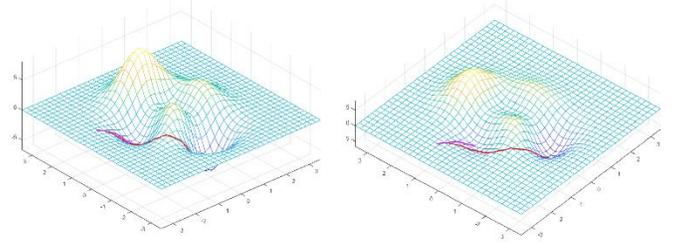

Figure 2 Schematic diagram of our modification on gradient descent.

Intuitively, our method can better prevent the agent from trapping in a local optima and learn faster with a greater convergence rate.

### C. Training Strategy

The actor-critic framework can be viewed as the traditional policy gradient method with a value network to estimate accumulated rewards. We first introduce the mathematical basis of policy gradient method to better clarify our approach. For each specific task, the actor which is approximated by a neural network with parameter $\theta$ will generate the distribution $\rho$ of action $a_t$ in correspondence to the input state $s_t$, where $t$ stands for time step. In practical use, we can either sample an action from $\rho$ or directly select the action with max probability. Then, the agent will receive a reward signal $r$ from the environment by performing the action, with which we can calculate the accumulated reward $R$ after each episode. Therefore, starting from initial state $s_0$, the agent constantly interacts with the environment, forming the sequence $\tau$: $\tau = \{s_0, a_0, r_0, s_1, a_1, r_1, \cdots, s_T, a_T, r_T\}$. The probability of the sequence $\tau$ occurring is addressed as $p_\theta(\tau)$. In order to update the policy, we need a function $J(\pi)$ which evaluates the utility of the policy. The function can be formulated as:

$$J(\pi) = \sum_\tau R(\tau) p_\theta(\tau) = E_{\tau \sim p_\theta(\tau)}[R(\tau)] = E_{\pi^{s_0}}[V(s_0)] \quad (3)$$

To maximize $J(\pi)$, we employ the gradient descent method to update the parameters of the network. The gradient of the objective function $J(\pi)$ is formulated as follows:

$$\begin{aligned}
\nabla_\theta J(\pi) &= \sum_\tau R(\tau) \nabla p_\theta(\tau) = \sum_\tau R(\tau) p_\theta(\tau) \frac{\nabla p_\theta(\tau)}{p_\theta(\tau)} \\
&= \sum_\tau R(\tau) p_\theta(\tau) \nabla log p_\theta(\tau) \\
&= E_{\tau \sim p_\theta(\tau)}[R(\tau) \nabla log p_\theta(\tau)] \\
&\approx \frac{1}{N} \sum_{n=1}^{N} R(\tau^n) \nabla log p_\theta(\tau^n) \\
&= \frac{1}{N \cdot T_n} \sum_{n=1}^{N} \sum_{t=1}^{T_n} R(\tau^n) \nabla log p_\theta(a_t^n | s_t^n) \quad (4)
\end{aligned}$$

Under the framework of actor-critic, we usually substitute the n-step reward $R(\tau^n)$ with Generalized Advantage Estimation (GAE) [24], which implies the action's advantage against others. The value of GAE is positive when the action is better than the average level in terms of Q-value, vice versa. $\nabla log p_\theta(a_t^n | s_t^n)$ signifies the gradient direction of the action's log-likelihood. Therefore, by performing gradient descent on $-J(\pi)$, we can increase the likelihood of actions that are above average and decrease those that are less promising. By merging $D_p$ with $E$ together to perform gradient descent, the gradient can be reformulated as:

$$\nabla_\theta J(\pi) = \frac{1}{N \cdot T_n} \sum_{n=1}^{N} \sum_{t=1}^{T_n} A(s_t^n, a_t^n) \nabla log p_\theta(a_t^n | s_t^n)$$
$$+ \frac{1}{N_{D_p}} \sum_{n=1}^{N_{D_p}} A(s^n, a^n) \nabla log p_\theta(a^n | s^n) \quad (5)$$

where $A(s_t^n, a_t^n) = Q^\pi(s_t^n, a_t^n) - V^\pi(s_t^n)$

With $D_p$ joining in, the log-likelihood of action following the demonstration guidance will increase even more. In other words, the agent takes a bigger step to descent loss contributed by $D_p$. Since the demonstration data stands for a better solution, our algorithm can thus manage to avoid local optima and reach near the global optima.

*D. Train Our Model*

We carry out our experiments mainly with on-policy PPO [1] algorithm based on AC framework, which is proved to obtain good results in many complex control problems. However, the methods we propose can be applied to all frameworks that engage the policy gradient method, such as SAC, TRPO, and A3C. The training procedure can be described sequentially as follows.

*1) Initialize the policy and value network with random parameters.*

*2) Collect a batch of sampled experience data $E$ by interacting with the environment and cluster them by observation and action.*

*3) Sample ratios of demonstration data $D$ and for each data frame $\{s_t^D, s_{t+1}^D, r_t^D, a_t^D\}$, verify if it resembles a certain cluster.*

*4) Randomly select a frame from the cluster if similar to demonstration, and restore them in $D_p$.*

*5) Repeat from step (2) for each new epoch until reaching preset epoch number to cease the process.*

*6) Explore and exploit until convergence*

## IV. EXPERIMENTS

*A. Results on standard benchmark environments*

To better verify the efficiency and effectiveness of our framework, we first carry out experiments on 4 OpenAI Gym environments with Mujoco [11], namely *Ant*, *Hopper*, *Humanoid* and *Swimmer*, as depicted in Fig. 1. Three of the environments involve solving a 2D locomotive control problem, while *Humanoid* is a 3D task where the robot tries to walk without tripping. Since *Humanoid* possesses many active joints, its dimension in the action space is much higher than the others which takes longer training. For each standard benchmark environment, the expert demonstration is obtained by replaying a SAC model which has already been trained to convergence, for 100 episodes each. In our experiments, we select the PPO algorithm to train from scratch and for each environment, we test our model on 5 different random seeds. Please refer to Table. 2 in *Appendix* for more hyperparameter details.

The learning curves of both our algorithm and PPO on each benchmark environment are shown in Fig. 3. Apparently, our algorithm can gain greater convergence rate than PPO during early and middle training stage. Especially in *Swimmer* and *Hopper*, we have remarkably improved the learning speed, with the guidance of the demonstration. Thanks to the fact that the dimension in action space is rather low in these two environments, which renders the demonstration data more dense and representative when evaluated with K-Means clustering metric, we can better leverage them to guide our agent. Moreover, greater final mean episode rewards are obtained in *Swimmer* with our method, which shows that the *know-how* is distilled from expert prior. As for *Ant* and *Humanoid*, although utilizing demonstration data has indeed equipped our agent with greater convergence rate during early training stage, the model exhibits over-fitting due to the sparsity of demonstration data, suffering from the curse of high dimensionality. However, this result did not undermine our algorithm from functioning in environments with high dimension in the action space, but point out the fact that a skillfully parameter handcrafting is imperative here.

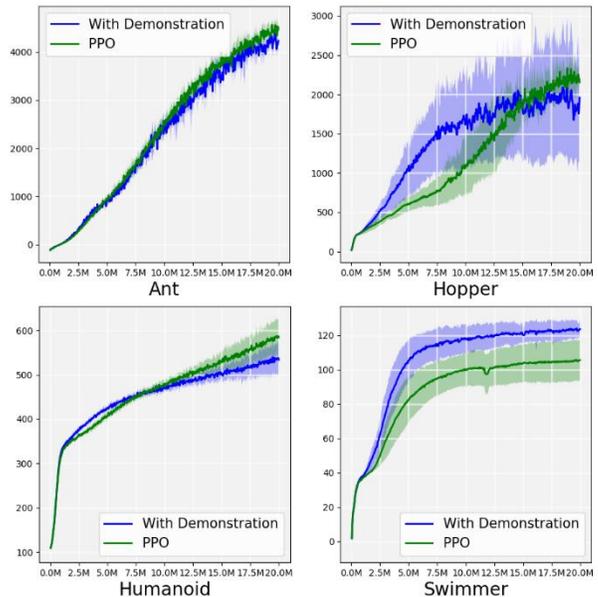

Figure 3 Learning curves of our algorithm and PPO on benchmark environments. X-axis describes the interaction steps; Y-axis shows the cumulative episode rewards.

Table. 1 is a quantification result of Fig. 3, which lists the ratio of time steps to reach the same mean episode rewards of our method with respect to PPO. Each row represents the steps needed for our method to reach $x\%$ of the maximum rewards that PPO gets. It can be deduced that we can improve the sample efficiency of our model in all 4 standard environments, thus obtaining a greater learning speed during certain stage of training. Although the final performance is weakened by over-fitting in *Ant* and *Humanoid*, we can still obtain a $10\% \sim 20\%$ improvement of sample efficiency.

TABLE 1  RATIO OF TIME STEPS TO REACH THE SAME MEAN EPISODE REWARDS.

|     | Ant  | Hopper | Humanoid | Swimmer |
|-----|------|--------|----------|---------|
| **85%** | 1.05 | 0.93   | 1.21     | 0.60    |
| **70%** | 1.04 | 0.63   | 0.91     | 0.71    |
| **50%** | 0.96 | 0.52   | 0.79     | 0.77    |
| **20%** | 0.91 | 0.64   | 1.00     | 1.00    |

Conclusively, our algorithm exhibits following properties:

*1) Increase sample efficiency, and especially the convergence rate during early training stage.*

*2) Avoid local optima, especially in low dimension environments, distilling knowledge from expert prior.*

*3) Heavily rely on parameter adjustment to prevent overfitting, and variance is introduced in our algorithm.*

### B. Results on our self-designed robots

To better apply our framework to real-life scenarios, we also carry out experiments in a simulated humanoid robot *InMoov* and a mechanical robot arm *Jaka*, as is shown in Fig. 4. The *InMoov* environment involves a life-size robot with 87 joints aiming to reach the tomato in the front. We partly re-designed the robot structure for human-robot synchronized control system [25], of which 53 joints can be actively controlled. Similarly, the *Jaka* environment consists of a task where the mechanical arm should press the button without contacting the obstacle. We develop the simulated version of the robot corresponding to its counterpart in real-life.

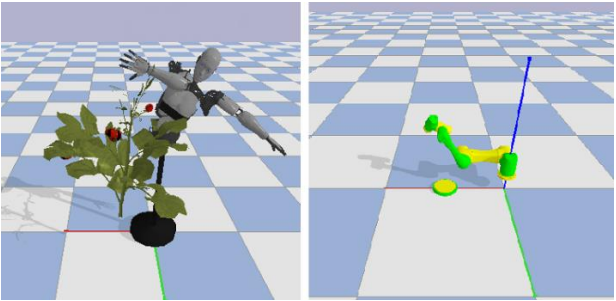

Figure 4 Self-designed robotic environments.

Both environments are complex with high dimension in the action and observation space. In *InMoov*, the reward signal depends on the distance between the end of robot's index finger and the target, while in *Jaka* a penalty when making contact with the obstacle is additionally added. In such a "learning to coordinate" task, the high dimensional inverse problem for robotic control is solved with reinforcement learning. Similarly, we develop our experiments with demonstration deriving from a well-trained PPO model. The learning curves are shown in Fig. 5.

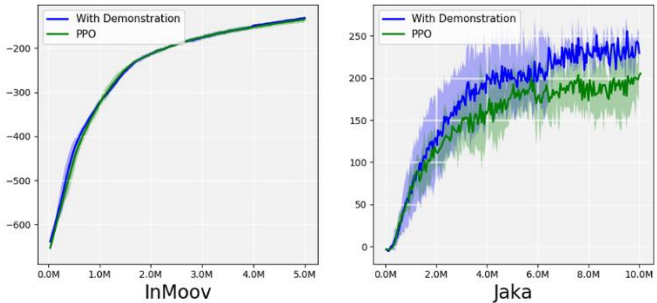

Figure 5 Learning curve of both algorithms in our self-designed robotic environments.

The results shown in Fig. 5 consolidate our conclusion in previous sections. Although the *InMoov* environment is high in dimension in the action space, the task is quite simple which involves finding a path that decreases the distance to target. That explains why the learning curve is smooth with little variance, since the optimal path is easy to find. The results in *Jaka* and *InMoov* verify that our framework could indeed accelerate convergence and reach higher final mean episode rewards, especially for complex non-linear robotic scenarios with high dimension in the action and observation space.

### C. Discussion

As we have noticed in Section *A* that our algorithm is prone to over-fit with inappropriate ratio of demonstration, especially in complex environments with high dimension in the action space, we thus conduct some ablation experiments on *Ant* and *Hopper*, further increasing the exploitation of demonstration data. The ratio of demonstration data is raised from 20% to 40%, and they are valid throughout the whole training process. The learning curves are illustrated in Fig. 6.

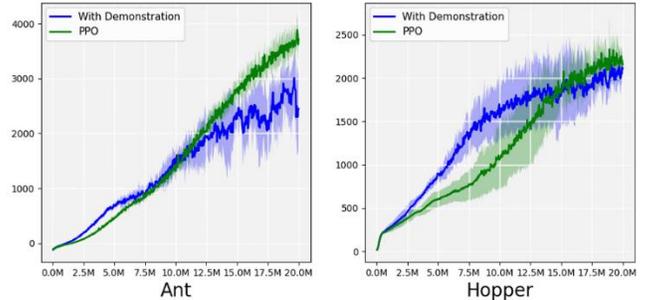

Figure 6 A depict of over-fitting of excessive exploitation of demonstration data.

Fig. 6 shows that the performance in *Ant* suffers from worse over-fitting than *Hopper*. Although the convergence rate is raised even more during early training stage, our model traps in local optima in both environments and oscillates drastically. Apparently, it is the excessive exploitation of demonstration data that causes even more severe over-fitting. Therefore, it is necessary to dynamically adjust the ratio of demonstration data. Fig. 3 and Fig. 6 also show that our algorithm is prone to introduce variance and may lower the stability of the system.

## V. CONCLUSION

In this study, we propose a novel framework for Learning from Demonstration (LfD) based on actor-critic algorithms. By leveraging the demonstration data to adjust gradient to update parameters, our LfD framework can better accustom to on-policy RL algorithms. By conducting experiments on 4 standard benchmark environments, we verify the property and advantage of our approach, which can boost convergence rate especially during early training stage and exhibit better final performance. However, our approach may degrade the stability of the system and relies on parameter adjustment to prevent over-fitting. Moreover, we also carry out experiments on 2 self-designed robotic environments and receive expected results. Therefore, it is proved that our framework is highly generalized and can be applied to solving complex control problems in real-life context. Our work is especially significant to complicated robotic scenarios where both action and observation space are in high dimension and agent-environment interactions are lengthy and expensive. This work is specially proposed for problems where only on-policy RL algorithms are available but explorations are somehow difficult to collect.

In terms of future work, we would like to find a more intuitive and convenient way to adjust the exploitation of demonstration, as well as a generic ratio for most problems which can improve convergence rate and final performance on the basis of maintaining stability. We would also like to deploy our framework with some other AC algorithms, to further verify the results.

### ACKNOWLEDGMENT

This work was supported by National Natural Science Foundation of China (Grant No.51775333), the Science Foundation of Shanghai Municipal Commission of Science and Technology (Grant No.18391901000) as well as the National Key Research and Development Project (2018YFB1305104).

### REFERENCES

[1] Schulman, J., Wolski, F., Dhariwal, P., Radford, A., & Klimov, O. (2017). Proximal policy optimization algorithms. arXiv preprint arXiv:1707.06347.
[2] Mnih, V., Badia, A. P., Mirza, M., Graves, A., Lillicrap, T., Harley, T., ... & Kavukcuoglu, K. (2016, June). Asynchronous methods for deep reinforcement learning. In International conference on machine learning (pp. 1928-1937). PMLR.
[3] Sutton, R. S., & Barto, A. G. (1998). Introduction to reinforcement learning (Vol. 135). Cambridge: MIT press.
[4] Gong, L., Sun, T., Li, X., Lin, K., Díaz-Rodríguez, N., Filliat, D., ... & Zhang, J. (2020). Demonstration Guided Actor-Critic Deep Reinforcement Learning for Fast Teaching of Robots in Dynamic Environments. IFAC-PapersOnLine, 53(5), 271-278.
[5] Argall, B. D., Chernova, S., Veloso, M., & Browning, B. (2009). A survey of robot learning from demonstration. Robotics and autonomous systems, 57(5), 469-483.
[6] Haarnoja, T., Zhou, A., Hartikainen, K., Tucker, G., Ha, S., Tan, J., ... & Levine, S. (2018). Soft actor-critic algorithms and applications. arXiv preprint arXiv:1812.05905.
[7] Lillicrap, T. P., Hunt, J. J., Pritzel, A., Heess, N., Erez, T., Tassa, Y., ... & Wierstra, D. (2015). Continuous control with deep reinforcement learning. arXiv preprint arXiv:1509.02971.
[8] Schaal, S. (1997). Learning from demonstration. Advances in neural information processing systems, 1040-1046.
[9] Konda, V. R., & Tsitsiklis, J. N. (2000). Actor-critic algorithms. In Advances in neural information processing systems (pp. 1008-1014).
[10] Rokach, L., & Maimon, O. (2005). Clustering methods. In Data mining and knowledge discovery handbook (pp. 321-352). Springer, Boston, MA.
[11] Todorov, E., Erez, T., & Tassa, Y. (2012, October). Mujoco: A physics engine for model-based control. In 2012 IEEE/RSJ International Conference on Intelligent Robots and Systems (pp. 5026-5033). IEEE.
[12] Parisi, S., Tateo, D., Hensel, M., D'Eramo, C., Peters, J., & Pajarinen, J. (2020). Long-term visitation value for deep exploration in sparse reward reinforcement learning. arXiv preprint arXiv:2001.00119.
[13] Hussein, A., Gaber, M. M., Elyan, E., & Jayne, C. (2017). Imitation learning: A survey of learning methods. ACM Computing Surveys (CSUR), 50(2), 1-35.
[14] Abbeel, P., & Ng, A. Y. (2004, July). Apprenticeship learning via inverse reinforcement learning. In Proceedings of the twenty-first international conference on Machine learning (p. 1).
[15] Cruz Jr, G. V., Du, Y., & Taylor, M. E. (2017). Pre-training neural networks with human demonstrations for deep reinforcement learning. arXiv preprint arXiv:1709.04083.
[16] Nair, A., McGrew, B., Andrychowicz, M., Zaremba, W., & Abbeel, P. (2018, May). Overcoming exploration in reinforcement learning with demonstrations. In 2018 IEEE International Conference on Robotics and Automation (ICRA) (pp. 6292-6299). IEEE.
[17] Vasan, G., & Pilarski, P. M. (2017, July). Learning from demonstration: Teaching a myoelectric prosthesis with an intact limb via reinforcement learning. In 2017 International Conference on Rehabilitation Robotics (ICORR) (pp. 1457-1464). IEEE.
[18] Hester, T., Vecerik, M., Pietquin, O., Lanctot, M., Schaul, T., Piot, B., ... & Gruslys, A. (2017). Learning from demonstrations for real world reinforcement learning.
[19] Vecerik, M., Hester, T., Scholz, J., Wang, F., Pietquin, O., Piot, B., ... & Riedmiller, M. (2017). Leveraging demonstrations for deep reinforcement learning on robotics problems with sparse rewards. arXiv preprint arXiv:1707.08817.
[20] Bellman, R. (1966). Dynamic programming. Science, 153(3731), 34-37.
[21] Lin, K., Gong, L., Li, X., Sun, T., Chen, B., Liu, C., ... & Zhang, J. (2020). Exploration-efficient deep reinforcement learning with demonstration guidance for robot control. arXiv preprint arXiv:2002.12089.
[22] Zuo, S., Wang, Z., Zhu, X., & Ou, Y. (2017, December). Continuous reinforcement learning from human demonstrations with integrated experience replay for autonomous driving. In 2017 IEEE International Conference on Robotics and Biomimetics (ROBIO) (pp. 2450-2455). IEEE.
[23] George, A. (2013). Efficient high dimension data clustering using constraint-partitioning k-means algorithm. Int. Arab J. Inf. Technol., 10(5), 467-476.
[24] Schulman, J., Moritz, P., Levine, S., Jordan, M., & Abbeel, P. (2015). High-dimensional continuous control using generalized advantage estimation. arXiv preprint arXiv:1506.02438.
[25] Gong, L., Gong, C., Ma, Z., Zhao, L., Wang, Z., Li, X., ... & Liu, C. (2017, August). Real-time human-in-the-loop remote control for a life-size traffic police robot with multiple augmented reality aided display terminals. In 2017 2nd International Conference on Advanced Robotics and Mechatronics (ICARM) (pp. 420-425). IEEE.

### APPENDIX

TABLE 2 HYPERPARAMETERS FOR THE PPO MODEL AND OUR FRAMEWORK WITH DEMONSTRATION

| Hyperparameter | Value |
| --- | --- |
| Policy learning rate | $3 \times 10^{-4}$ |
| Value function learning rate | $10^{-3}$ |
| $\gamma$ | 0.99 |
| $\lambda$ | 0.97 |
| Clip ratio ($\epsilon$) | 0.2 |
| Target KL to clip | 0.01 |
| Update interval | 40 |